\documentclass{llncs}
\usepackage[english]{babel}
\usepackage[utf8x]{inputenc}
\usepackage[T1]{fontenc}
\usepackage{etoolbox}
\usepackage{lipsum}

\usepackage{amsmath}
\usepackage{graphicx}
\usepackage[colorinlistoftodos]{todonotes}
\usepackage[colorlinks=true, allcolors=blue]{hyperref}

\makeatletter
\renewcommand\subsubsection{\@startsection{subsubsection}{3}{\z@}%
                                     {-3.25ex\@plus -1ex \@minus .2ex}%
                                     {1.5ex \@plus .2ex}%
                                     {\normalfont\normalsize\bfseries}}
\makeatother

\begin{document}

\title{CORE-GPT: Combining Open Access research and large language models for credible, trustworthy question answering. }

%
\titlerunning{CORE-GPT}  
%
\author{David Pride, Matteo Cancellieri \and Petr Knoth}
\authorrunning{Pride et al.} 
%
\tocauthor{}
\institute{The Knowledge Media Institute, The Open University, Milton Keynes, UK.\\
\email{\{david.pride, matteo.cancellieri, petr.knoth\}@open.ac.uk}}

\maketitle

\begin{abstract}

In this paper, we present CORE-GPT, a novel question-answering platform that combines GPT-based language models and more than 32 million full-text open access scientific articles from CORE\footnote{https://core.ac.uk}. We first demonstrate that GPT3.5 and GPT4 cannot be relied upon to provide references or citations for generated text. We then introduce CORE-GPT which delivers evidence-based answers to questions, along with citations and links to the cited papers, greatly increasing the trustworthiness of the answers and reducing the risk of hallucinations. CORE-GPT's performance was evaluated on a dataset of 100 questions covering the top 20 scientific domains in CORE, resulting in 100 answers and links to 500 relevant articles. The quality of the provided answers and and relevance of the links were assessed by two annotators. Our results demonstrate that CORE-GPT can produce comprehensive and trustworthy answers across the majority of scientific domains, complete with links to genuine, relevant scientific articles.

\end{abstract}

\section{Introduction}
LLMs demonstrate a remarkable ability to process and interpret natural language, understanding various nuances and intricacies of human language. They excel at text generation, crafting coherent and contextually relevant responses or content, ranging from casual conversations to technical articles. However, these are predictive models and cannot be relied upon to provided reliable sources or citations for any generated text.

\begin{figure}[!h]
  \centering
  \hspace*{0.25cm}\includegraphics[width=0.8\linewidth]{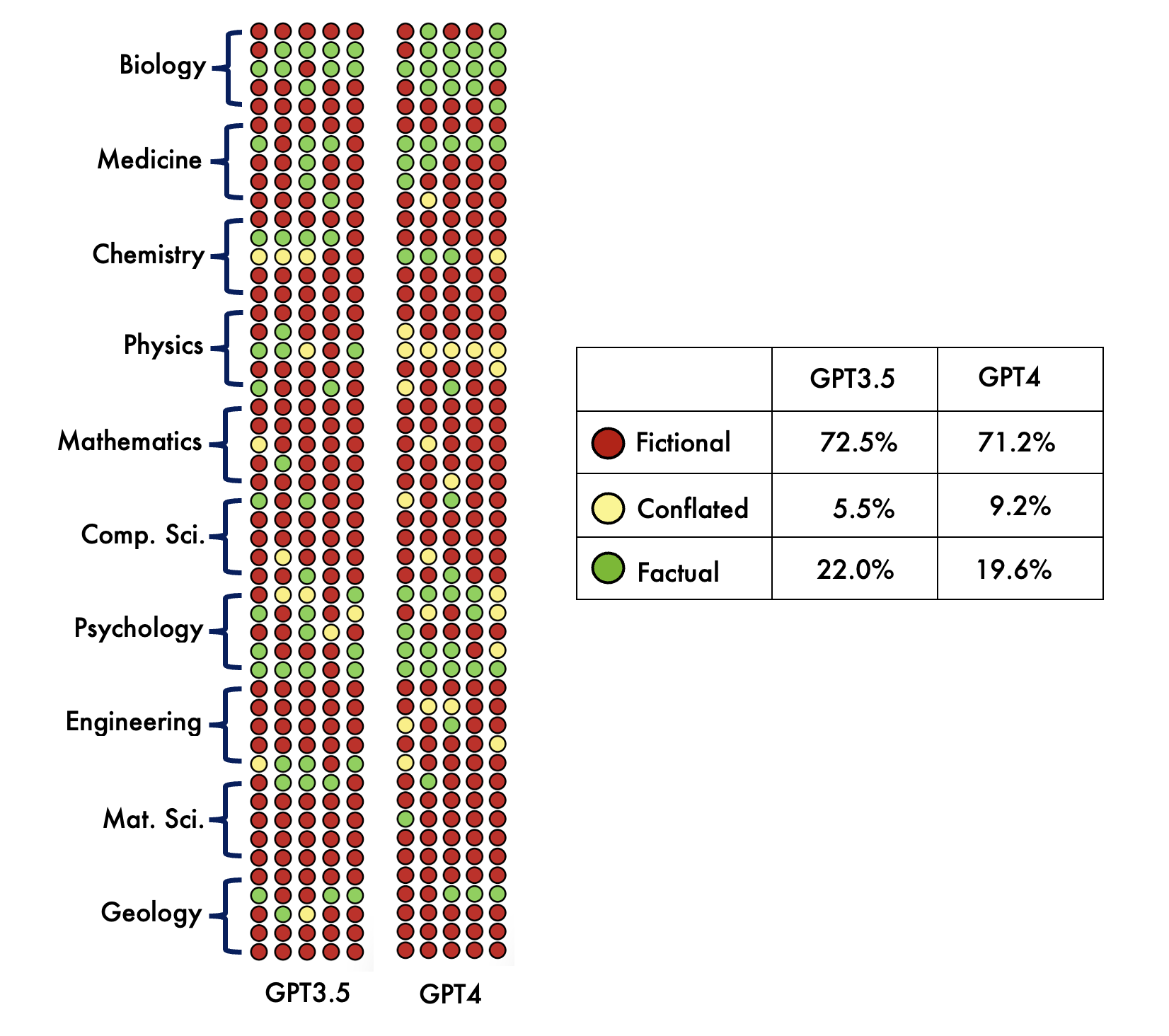}
  \caption{Citations to answers given by LLMs. Each row represents 5 sources / citations for a single answer. Overall, 72.5\% of citations provided by GPT3.5 were fictional. This figure was 71.2\% for GPT4}
  \label{fig:chat-gpt-fact-fiction}
\vspace{-15pt}
\end{figure}

In order to better understand the problem, we used the GPT3.5 and GPT4 models to answer 50 questions from across ten different domains, and to provide the five top sources / citations for each of the answers. Each row in Figure \ref{fig:chat-gpt-fact-fiction} shows the results for a single answer. A green dot represents a genuine, factual citation with a paper that exists or a link that goes directly to the paper itself. A red dot represents a completely fictional paper that simply does not exist. The yellow dots were used where there was what we termed \textit{conflation}, meaning the provided citation or source was not real, but used either a mix of real titles or real author names or then linked to a completely different paper entirely. This shows that 22\% of references for GPT3.5 and less than 20\% for GPT4 were factual. 

\noindent Whilst it can be argued that GPT3.5 and GPT4 were not designed to reference evidence \cite{lse2022ai}, it can be widely observed that people have attempted to use them for these purposes and that it would be valuable if they could be used in this way. In this paper, we address this issue by introducing CORE-GPT. Our main contributions are:
\vspace{-5pt}
\begin{itemize}
    \item We provided empirical evidence demonstrating that GPT3.5 and GPT4 cannot be relied upon to generate sources of references. 
    \item We provide a solution that combines the power of GPT models and a global open research corpus to deliver a credible and trustworthy question-answering solution, accompanied with references to research literature. 
    \item Our question-answering solution is capable of providing answers including references to recently published research without the need for retraining the GPT models.     
\end{itemize}

\vspace{-10pt}
\section{Related work}

The term Large Language Model has been in existence for many decades, however the LLMs we focus on here are extensions of the \textit{transformer} model architecture introduced in 2017 by Vaswani et al. in their seminal paper \textit{'All you need is attention'} which lead to the development of the BERT transfomer models and its siblings SciBERT \cite{beltagy2019scibert} and RoBERTa \cite{liu2019roberta}) and to GPT-2 \cite{radford2019language},3 \cite{brown2020language} and most recently GPT4 \cite{gpt4}. The advancements and overall recent developments in LLMs have been exhaustively reviewed by several scholars, including Fan et al. \cite{fan2023bibliometric} and Zhao et al. \cite{zhao2023survey}, whose comprehensive surveys offer in-depth analyses of this rapidly evolving discipline. This paper will therefore not reiterate these developments. 

LLMs have demonstrated remarkable capabilities in many areas. There are however significant challenges associated with the use of LLMs. There has been concerns about the risk of plagiarism and the potential impact on education and assessment \cite{kasneci2023chatgpt}. There are also specific concerns about the implications for the medical \cite{shen2023chatgpt} and legal \cite{BBC2023} domains. Beyond these domain-specific concerns, the robustness of LLMs has also been questioned. Issues such as  hallucinations, or the generation of statements that appear credible but are in fact entirely fabricated, have been widely reported. 

In a study of particular interest to scientists and researchers, Gao et al. \cite{gao2022comparing} showed that models based on the Generative Pre-training Transformer (GPT) architecture could generate abstracts for scientific articles that were often indistinguishable from those authored by humans. However, Alkaissi and McFarlane \cite{alkaissi2023artificial} conducted a study to evaluate ChatGPT's ability to answer complex biomedical and healthcare-related questions. Their results demonstrated mixed quality in the generated responses, with answers consisting of a blend of factual information and fabricated content. Crucially, when ChatGPT was asked to provide citations and PubMed IDs to support its answers, all the provided references were fictional, and the given PubMed IDs were simply sequences of random numbers with no relation to existing papers. This research, corroborated by additional studies \cite{smu}, underscores a profound problem with LLMs generating authentic-sounding but entirely fictional content.

These challenges and the results shown in Table \ref{fig:chat-gpt-fact-fiction} highlight a significant hurdle that needs to be overcome in order to be able to leverage the abilities of LLMs for question answering whilst limiting the potential for false or misleading answers.. The focus of our work in this paper is on addressing this credibility gap, by proposing a novel approach that combines Open Access scientific literature with LLMs to enhance the reliability and trustworthiness of these systems. 

\section{Our solution - CORE-GPT}

\subsection{CORE-GPT Workflow}

CORE-GPT has been developed specifically to address the problems discussed in the previous sections. We use a three-stage approach to returning answers to user questions with links to relevant full-text papers in CORE.

\begin{figure}[]
\vspace{-10pt}
  \centering
  \includegraphics[width=0.9\linewidth]{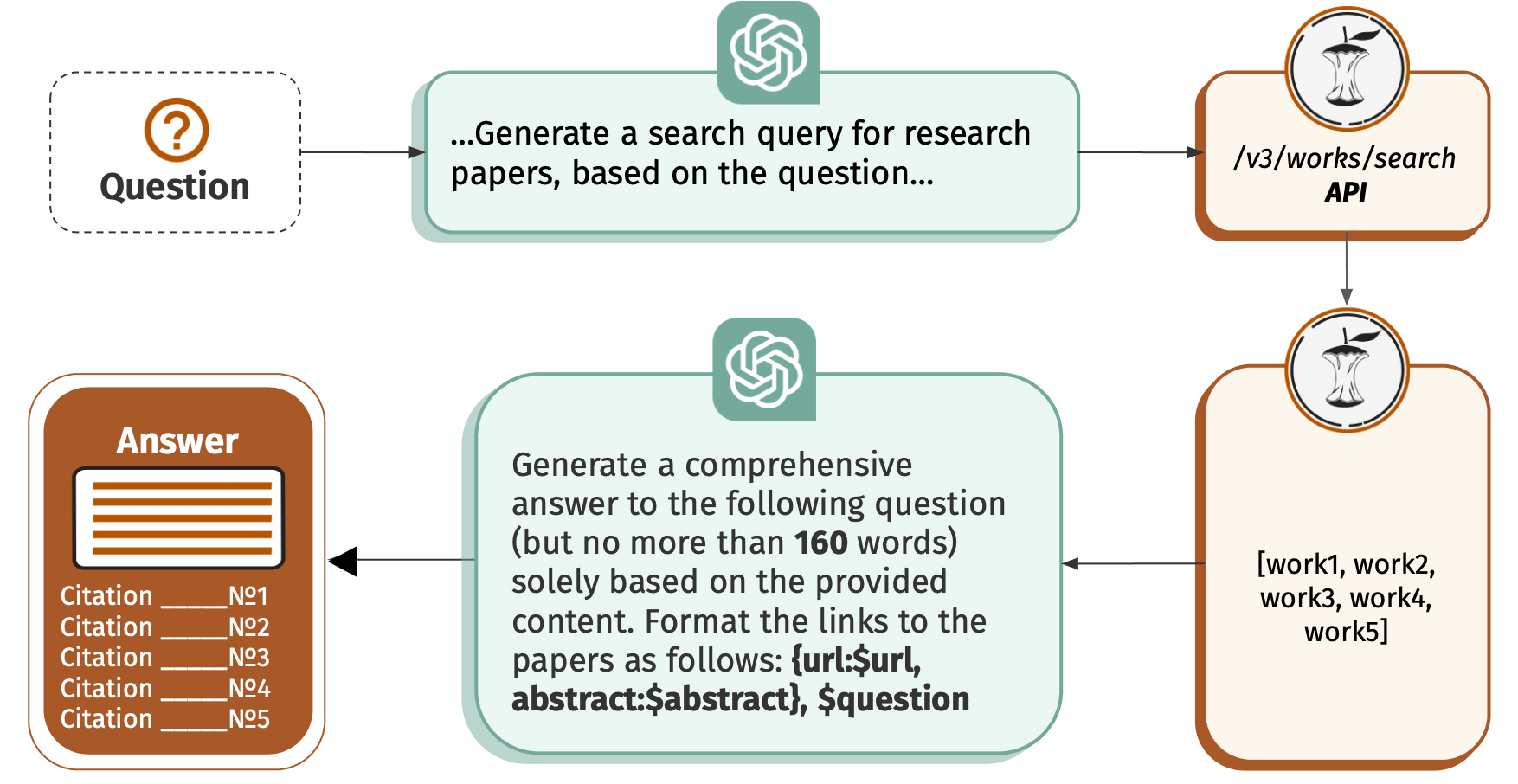}
  \caption{CORE-GPT workflow.}
  \label{fig:subjects}
\vspace{-10pt}
\end{figure}

\noindent In Stage 1, the original question is passed to the GPT4 API with several instructions. 

\begin{itemize}
  \item Identify the key terms within the question
  \item Enrich with close synonyms
  \item Formulate this into a search query. 
\end{itemize}

\noindent A sample question and search formatted response can be seen below: 

\begin{quote}
    \textbf{Original user question}\newline
    \textit{What strategies can be implemented to improve literacy rates in rural primary schools in developing countries?}\newline

    \textbf{Formatted query}\newline
    \textit{strategies improve literacy rates rural primary schools developing countries OR low-income OR underdeveloped OR third-world}
\end{quote}

\noindent In Stage 2, the formatted search query is then passed to the CORE API which returns the five most relevant papers where the full-text content is available. Stage 3 is the key to the novel solution provided by CORE-GPT. We pass the titles and abstracts returned in Stage 2 back to the GPT4 API with further instructions:

\begin{quote}
    \textit{Generate a comprehensive answer to the following question (but no more than 160 words) solely based on the content provided. Format the links to the papers as follows: furl:Surl, abstract:\$abstract}, \$question
\end{quote}

\noindent Our evaluation shows that this critical third stage is largely effective at constraining the model to base its reply only on the supplied input. The answer and provided links are then shown to the user. The full workflow can be seen in Figure \ref{fig:subjects}

\begin{figure}[!ht]
\centering
  \frame{\includegraphics[width=0.7\linewidth]{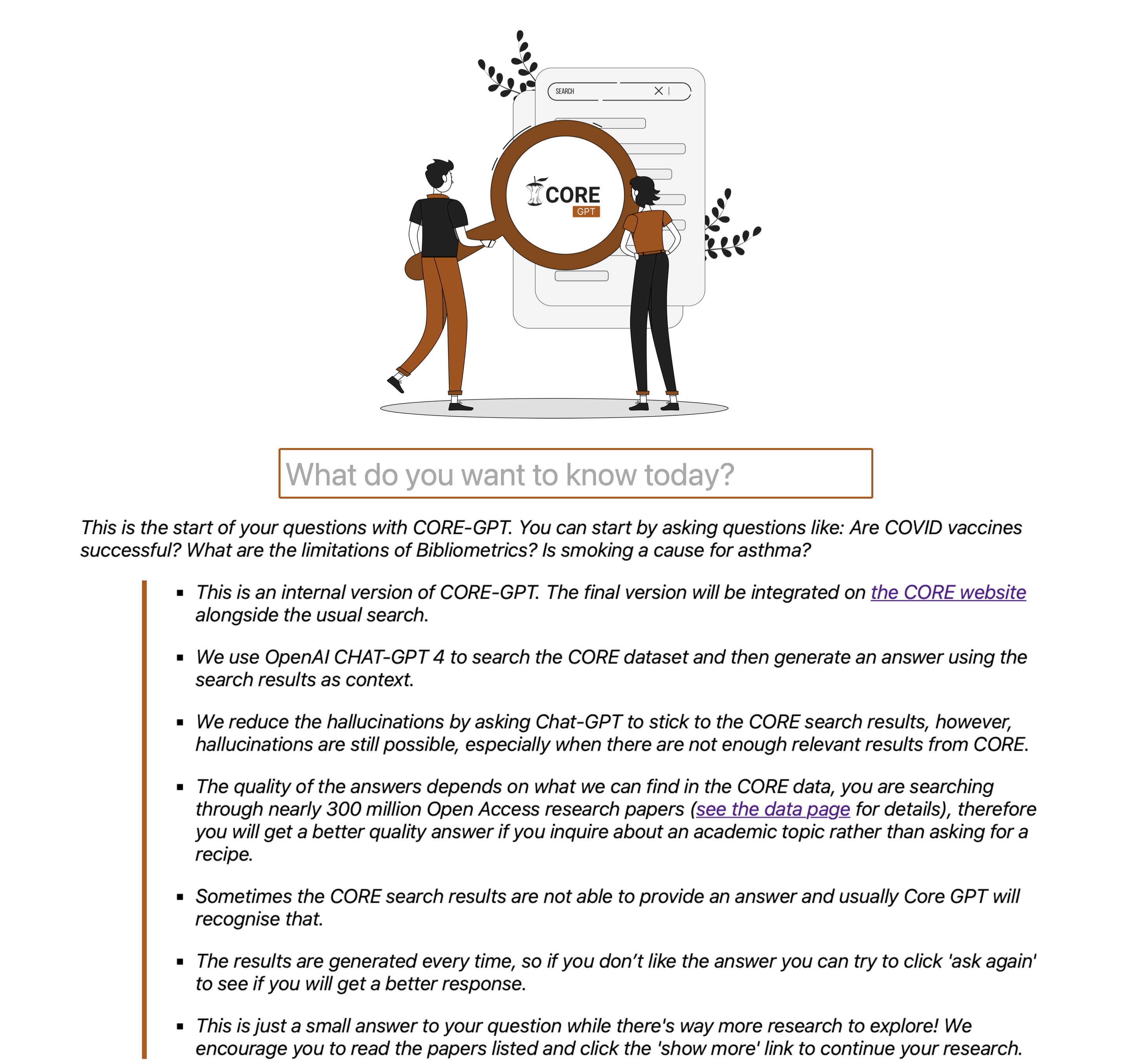}}
  \caption{CORE-GPT user interface.}
  \label{fig:ui}
\vspace{-20pt}
\end{figure}

\subsection{The CORE-GPT User Interface}

\begin{figure}[!h]
  \frame{\includegraphics[width=\linewidth]{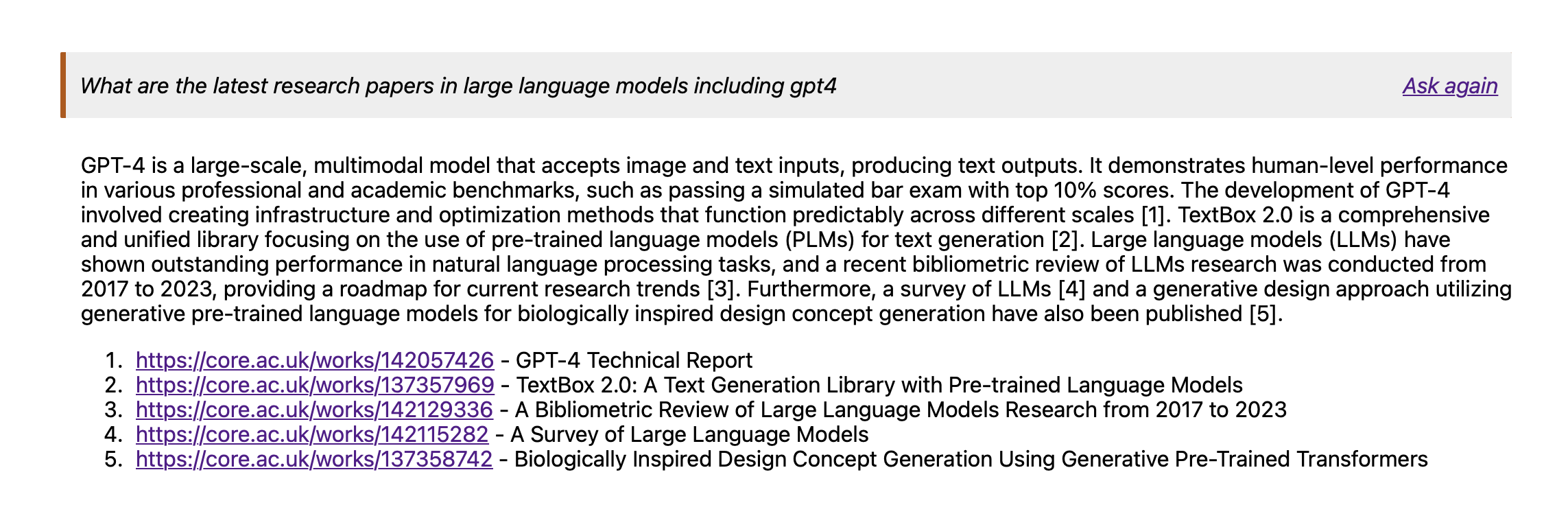}}
  \caption{CORE-GPT Sample results including very recently published papers (less than one month since publication.)}
  \label{fig:results}
\vspace{-20pt}
\end{figure}

Initially, CORE-GPT will be made available on the CORE website as a new web-based question / answer platform.(Figure \ref{fig:ui}). Further development will allow for the service to be made available via the CORE API. This is discussed in the Future Work section (Section \ref{future}.) A sample result is shown in Figure \ref{fig:results}.

\subsection{Benefits of CORE-GPT}

The key benefit of CORE-GPT is in ensuring that the content of the generated answers is drawn from published scientific literature, which is then subsequently referenced. This greatly reduces the potential for hallucinations. There are further benefits derived from the constraints placed on the model. In our evaluation, there were instances where, despite the massive-scale corpus that CORE-GPT draws its answers from, there was not enough relevant content to formulate a comprehensive answer. Below is an example question from the questions dataset used for the evaluation were this was the case;

\begin{quote}
    "What are the potential long-term health impacts of regular use of over-the-counter pain medications on the liver and kidney function in young adults?"
\end{quote}
\vspace{-5pt}
\noindent In cases like these, the GPT4 model is capable of recognising the lack of relevant responses. If a complete answer cannot be given, the user will be informed with the following type of message;

\begin{quote}
    "Regular use of over-the-counter (OTC) pain medications can potentially impact liver and kidney function in young adults. \textit{However, the provided results do not offer specific information} on the long-term health impacts of such medications on these organs. \textit{To obtain a comprehensive answer, further research on this topic would be necessary}."
\end{quote}

\noindent In our evaluation we found that whilst this type of answer was understandably low scoring in terms of comprehensiveness and utility, it scored highly for trustworthiness. The key factor here is that the model is forced to be honest when it does not know something. This greatly reduces the potential for hallucinations and increases the overall viability and usability of CORE-GPT in academic question / answering. 

Another key benefit is intrinsically linked to the way CORE operates as an Open Access infrastructure. Anyone who has used the latest GPT models will almost certainly be familiar with the response \textit{'I'm sorry for the inconvenience. As an AI model with a knowledge cutoff in September 2021, I don't have real-time information'}. CORE however is constantly aggregating content from the global network of Open Access repositories and as soon as a document is indexed in CORE, it is available to CORE-GPT to be used in answers and cited. The search shown in Figure \ref{fig:results} was undertaken during the second week of May 2023. The results contain papers published as recently as April 2023. As CORE-GPT is designed to work in this way, this removes the knowledge cut-off date experienced when using just the GPT models themselves.

\section{Evaluation Methodology}

\subsection{Data Sources}

CORE-GPT is designed to provide citations to the research papers used to formulate the answers. All cited research papers are drawn from the CORE corpus of Open Access research literature. CORE is currently the one of the largest aggregators of OA scholarly knowledge, aggregating content from the global network of almost 11,000 institutional, pre-print and publisher repositories and, as of May 2023, hosts over 32 million full-text research papers. \cite{naturedata}

\begin{table}
\begin{centering}
\begin{tabular}{|l|r|}
    \hline 
    Metadata records & 291,151,257 \\
    \hline
    Records with full text & 32,812,252 \\
    \hline
    Records with abstract & 94,521,867 \\
    \hline
    Records with full-text link & 139,000,000\textsuperscript{\dag} \\
    \hline
    Data providers & 10,744 \\ 
    \hline
    Number of CORE data provider countries & 150 \\ 
    \hline
    Estimated number of languages of collected content & 118 \\
    \hline
\end{tabular}
\caption{Size of the CORE collection as of January 2023. \textsuperscript{\dag}Estimate based on analysis}
\label{tab:core_size}
\end{centering}
\vspace{-10pt}
\end{table}
\subsection{Question generation}

Our first task was to generate a dataset of questions that could be used to test the performance of CORE-GPT and also to compare this performance against large language models such as GPT3.5 and GPT4. Additionally, we wanted to ascertain whether the models themselves and also CORE-GPT were more successful in some domains and less successful in others. We therefore generated a dataset of questions based on the split of domains in the CORE dataset. The domains with the largest amount of full text content in CORE were selected. We added education as the final domain to give 20 domains. 

\begin{figure}[!h]
\vspace{-20pt}
  \includegraphics[width=0.9\linewidth]{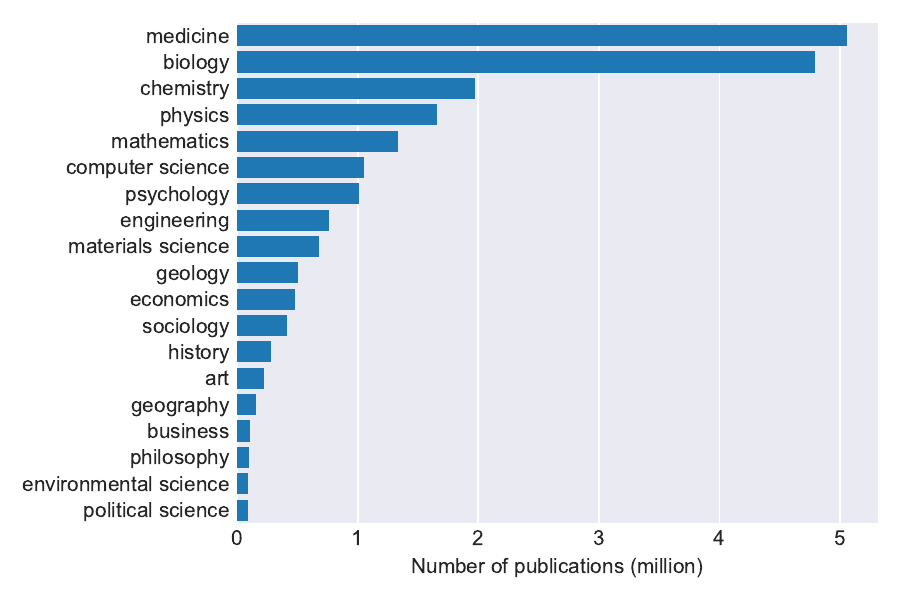}
  \caption{Subject distribution of a sample of 20,758,666 CORE publications.}
  \label{fig:core-subjects}
\end{figure}

To aid in the rapid development of the questions dataset, we elected to use a large language model. GPT-4 was chosen for its recency and known abilities for this task. Using the list of domains previously discussed, the OpenAI GPT-4 API was used to generate the questions using the following prompt; 

\begin{quote}
            messages=[
            {"role": "system", "content": "\textit{write a graduate level research question in the following domain, only reply with the body of the question itself}:"},
            {"role": "user", "content": \textit{domain}},
        ]
\end{quote}

\noindent Five questions were generated from each domain for a total of 100 questions. Overall, the question generation methodology was effective and allowed for rapid generation of the questions dataset. There are however some potential limitations that this method may introduce which are discussed in the Discussion section (Section \ref{discussion}.) The datasets of all questions and answers with accompanying citations can be found in the Github repository for this study\footnote{https://github.com/oacore/core-gpt-evaluation}.
\vspace{-5pt}
\subsection{Evaluation Metrics}

Effectively evaluating CORE-GPT requires a two-step approach as both the given answer and the provided citations must be validated. We elected to use three metrics for each of the answers as follows: 

\begin{itemize}
  \item \textbf{Comprehensive:}
 How comprehensively is the question answered?  
 \item \textbf{Trust:} How trustworthy is the answer?
  \item \textbf{Utility:} How useful is the answer?
\end{itemize}

\noindent For the citations, we use \textbf{relevance} as the metric, that is how relevant is the given reference to the original question. To enable evaluation of the results, a browser-based evaluation platform was developed which sequentially displayed each of the 100 questions and answers and the title, abstracts and links to the five papers for each answer. For each question, the three answer metrics shown above and the relevance score for each of the citations could be assigned a value from zero to ten.

\begin{table}[]
\vspace{-10pt}
\centering
\begin{tabular}{|l|l|}
\hline
\textbf{Class}    & \textit{\textbf{Agreement (k)}} \\ \hline
Comprehensiveness & 0.792                           \\ \hline
Trust             & 0.760                           \\ \hline
Utility           & 0.748                           \\ \hline
Cite 1            & 0.808                           \\ \hline
Cite 2            & 0.727                           \\ \hline
Cite 3            & 0.665                           \\ \hline
Cite 4            & 0.717                           \\ \hline
Cite 5            & 0.651                           \\ \hline
\end{tabular}
\caption{Inter-annotator agreement for each classification}
\
\label{tab:inter_a}
\vspace{-25pt}
\end{table}
\vspace{-5pt}
\noindent Two annotators were retained and were given written instructions and training using the evaluation platform with sample data. Inter-annotator agreement for each metric was measured using Cohen's Kappa with quadratic weights. This measure was chosen for the task as it accounts for both small and large differences of opinion more accurately than unweighted Kappa. The results for the inter-annotator agreement can be seen in Table \ref{tab:inter_a}. 

\section{Results}

\vspace{-5pt}
\subsection{Quality of answers}
\par

\begin{figure}[!h]
\vspace{-10pt}
  \centering
  \includegraphics[width=0.9\linewidth]{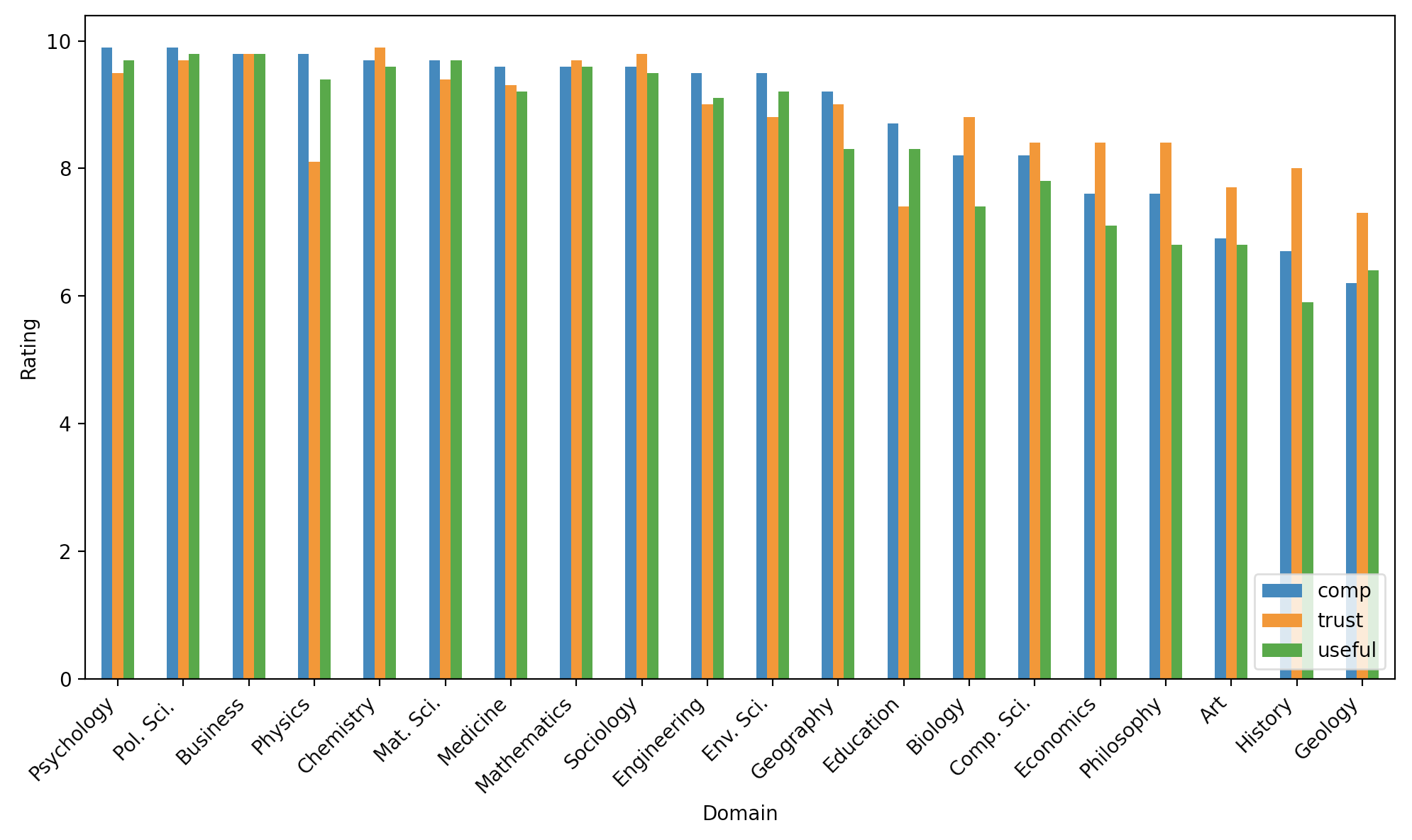}
  \caption{Mean comprehensiveness, trust and utility scores for each domain ordered by mean comprehensiveness.)}
  \label{fig:domain_means}
\vspace{-10pt}
\end{figure}

Using the evaluation platform, the annotators were asked to rank each answer according to the three metrics introduced previously, \textit{comprehensiveness}, \textit{trust} and \textit{utility}. Each of these metrics could be scored from 0 (not at all) to 10 (completely) for each answer. Figure \ref{fig:domain_means} shows the mean comprehensiveness, trust and utility scores for the answers from each of the 20 domains. CORE-GPT performs exceptionally well across most domains, but is less successful in a few areas.  In 75\% of the domains, the mean comprehensive, trust and utility score was 8 points or greater, and 9 points or greater in over half of the domains, indicating that CORE-GPT provides highly relevant, factual and, most importantly, referenced answers. A full breakdown of all scores is shown in Tables \ref{tab:tabCTU} and \ref{tab:tabCITE}. It is worth noting that in the domains where the answers were deemed by the annotators to be less comprehensive and less useful, the trust scores remained fairly high (>6 across all domains) indicating that overall the given answers were considered trustworthy. 

\vspace{-5pt}


We investigated whether there was a relationship between the domain scores for comprehensiveness, trustworthiness and utility and the number of research papers in CORE for each respective domain (Figure \ref{fig:core-subjects}).  However, we found only a weak correlation (Pearson's R0.23, n=20), indicating that having less research content in some domains does not fully explain the lower performance in these areas. CORE is a comprehensive source of multidisciplinary research content \cite{gusenbauer2022search} and it might be that the domains in which there is genuinely less content are not necessarily insufficiently represented in CORE.  

We further examined whether the length of the abstracts given to the model to generate the answers had an impact on the quality scores for the answers. There is a wide variance in mean abstract length across the domains, from economics (171 words) to engineering (329 words), we were therefore interested to see if this influenced the scores for comprehensiveness and utility. However, we observed no correlation  between these scores and the mean abstract lengths in each domain.(Pearson's $r=-0.02, n=20$)



\subsection{Citation Relevance}

\begin{figure}[!h]
\vspace{-5pt}
  \centering
  \includegraphics[width=0.9\linewidth]{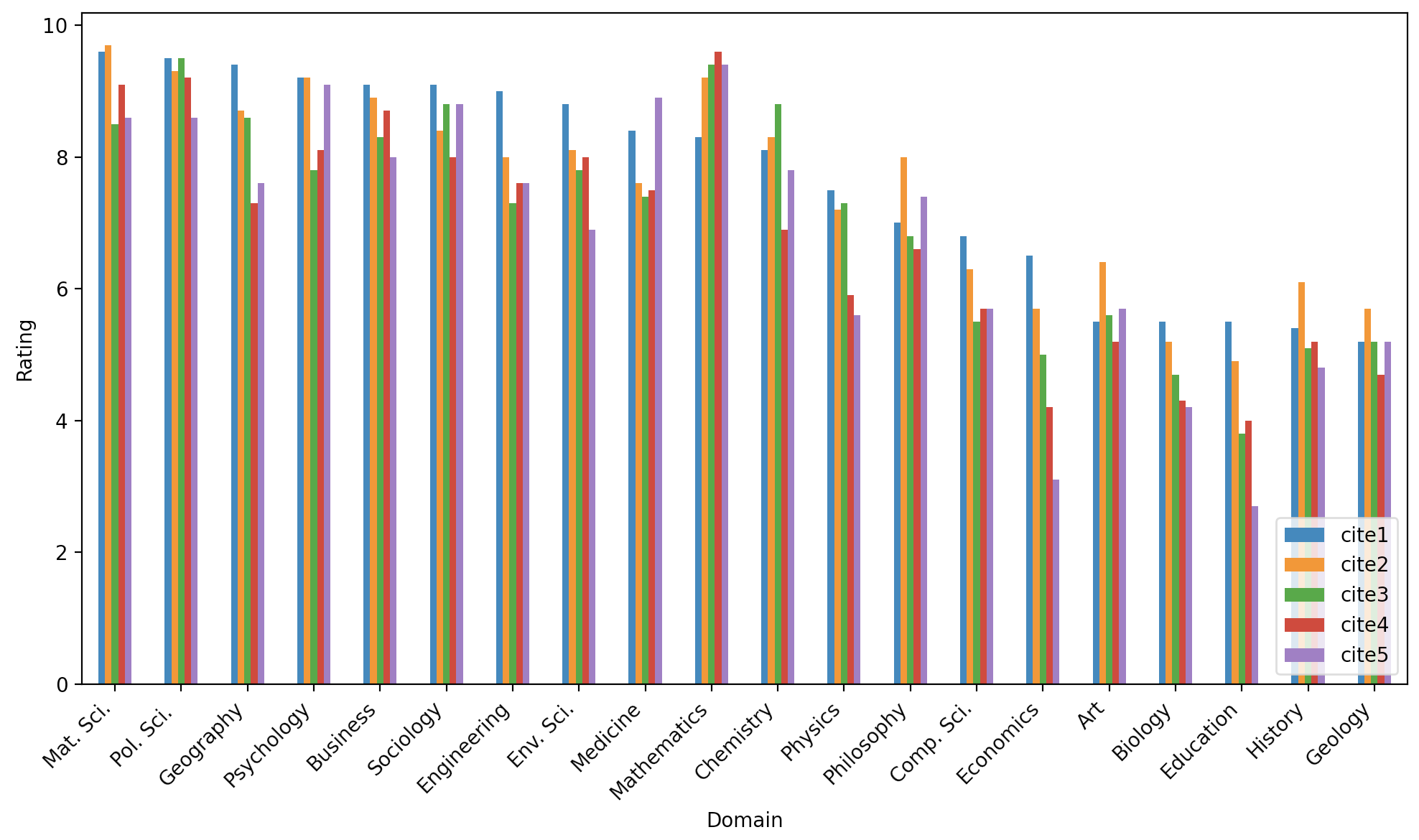}
  \caption{Mean citation relevance scores for each domain. (Ordered by relevance score for first citation.)}
  \label{fig:domain_cites_means}
\vspace{-20pt}
\end{figure}

In contrast to the results for GPT3.5 and GPT4 shown in Figure \ref{fig:chat-gpt-fact-fiction}, all citations provided by CORE-GPT are, by design, links to genuine research papers. Therefore the evaluation was based on testing not the existence of these papers, but their relevance to the user's original question. The annotators were asked to rank each citation from 0 (not relevant at all) to 10 (completely relevant). Figure \ref{fig:domain_cites_means} shows the mean relevance score for each of the five citations across all domains.

Based on the previously discussed Figure \ref{fig:domain_means} we observed that CORE-GPT provides comprehensive, trustworthy and useful answers for the majority of the domains. However, in some domains, such Geology, History and Art, comprehensiveness and utility were lower. We were therefore interested to find out to what extent the ability of CORE-GPT to provide good-quality answers is linked to the quality of the retrieved references. We found that there is a very strong correlation between the relevance of the retrieved references and comprehensiveness, trust and utility across domains respectively (Pearson $r = 0.77$ (comp.); $r = 0.83$ (trust); $r = 0.80$ (utility), $n=20$). This suggests that the ability to retrieve relevant references with respect to a user's question has a major impact on the quality of CORE-GPT's answers.  

The annotators were asked to score the relevance of each of the five retrieved references separately, enabling us to test the performance of our reference retrieval functionality.  A well optimised ranking function should retrieve the most relevant references first. As a result, we expected to observe that the top retrieved references would be assigned higher relevance scores than the latter references by the annotators on average. The results reported in Table \ref{tab:tabCITE} indeed confirms this trend. 
\vspace{-15pt}
\begin{table}[!h]
\hspace*{-1.3cm}
\begin{minipage}{.5\linewidth}
\centering
\begin{tabular}{|l|l|l|l|l|}
\hline
\textbf{Domain} & \textbf{Comp} & \textbf{Trust} & \textbf{Utility} & \textbf{Mean} \\ \hline
\textbf{Pol. Sci.} & 9.9 & 9.7 & 9.8 & 9.80 \\ \hline
\textbf{Business} & 9.8 & 9.8 & 9.8 & 9.80 \\ \hline
\textbf{Chemistry} & 9.7 & 9.9 & 9.6 & 9.73 \\ \hline
\textbf{Psychology} & 9.9 & 9.5 & 9.7 & 9.70 \\ \hline
\textbf{Mathematics} & 9.6 & 9.7 & 9.6 & 9.63 \\ \hline
\textbf{Sociology} & 9.6 & 9.8 & 9.5 & 9.63 \\ \hline
\textbf{Mat. Sci.} & 9.7 & 9.4 & 9.7 & 9.60 \\ \hline
\textbf{Medicine} & 9.6 & 9.3 & 9.2 & 9.37 \\ \hline
\textbf{Engineering} & 9.5 & 9 & 9.1 & 9.20 \\ \hline
\textbf{Env. Sci.} & 9.5 & 8.8 & 9.2 & 9.17 \\ \hline
\textbf{Physics} & 9.8 & 8.1 & 9.4 & 9.10 \\ \hline
\textbf{Geography} & 9.2 & 9.0 & 8.3 & 8.83 \\ \hline
\textbf{Education} & 8.7 & 7.4 & 8.3 & 8.13 \\ \hline
\textbf{Comp. Sci.} & 8.2 & 8.4 & 7.8 & 8.13 \\ \hline
\textbf{Biology} & 8.2 & 8.8 & 7.4 & 8.13 \\ \hline
\textbf{Economics} & 7.6 & 8.4 & 7.1 & 7.70 \\ \hline
\textbf{Philosophy} & 7.6 & 8.4 & 6.8 & 7.60 \\ \hline
\textbf{Art} & 6.9 & 7.7 & 6.8 & 7.13 \\ \hline
\textbf{History} & 6.7 & 8 & 5.9 & 6.87 \\ \hline
\textbf{Geology} & 6.2 & 7.3 & 6.4 & 6.63 \\ \hline
\end{tabular}
\caption{Mean answer quality scores for all domains.}
\label{tab:tabCTU}
\end{minipage}%
\hspace*{1cm}
\begin{minipage}{.5\linewidth}
\centering
\begin{tabular}{|l|l|l|l|l|l|l|}
\hline
\textbf{Domain} & \multicolumn{1}{r|}{\textbf{cite1}} & \multicolumn{1}{r|}{\textbf{cite2}} & \multicolumn{1}{r|}{\textbf{cite3}} & \multicolumn{1}{r|}{\textbf{cite4}} & \multicolumn{1}{r|}{\textbf{cite5}} & \multicolumn{1}{r|}{\textbf{Mean}} \\ \hline
\textbf{Pol. Sci.} & 9.5 & 9.3 & 9.5 & 9.2 & 8.6 & 9.22 \\ \hline
\textbf{Mathematics} & 8.3 & 9.2 & 9.4 & 9.6 & 9.4 & 9.18 \\ \hline
\textbf{Mat. Sci.} & 9.6 & 9.7 & 8.5 & 9.1 & 8.6 & 9.10 \\ \hline
\textbf{Psychology} & 9.2 & 9.2 & 7.8 & 8.1 & 9.1 & 8.68 \\ \hline
\textbf{Sociology} & 9.1 & 8.4 & 8.8 & 8 & 8.8 & 8.62 \\ \hline
\textbf{Business} & 9.1 & 8.9 & 8.3 & 8.7 & 8 & 8.60 \\ \hline
\textbf{Geography} & 9.4 & 8.7 & 8.6 & 7.3 & 7.6 & 8.32 \\ \hline
\textbf{Chemistry} & 8.1 & 8.3 & 8.8 & 6.9 & 7.8 & 7.98 \\ \hline
\textbf{Medicine} & 8.4 & 7.6 & 7.4 & 7.5 & 8.9 & 7.95 \\ \hline
\textbf{Env. Sci.} & 8.8 & 8.1 & 7.8 & 8 & 6.9 & 7.92 \\ \hline
\textbf{Engineering} & 9 & 8 & 7.3 & 7.6 & 7.6 & 7.90 \\ \hline
\textbf{Philosophy} & 7 & 8 & 6.8 & 6.6 & 7.4 & 7.16 \\ \hline
\textbf{Physics} & 7.5 & 7.2 & 7.3 & 5.9 & 5.6 & 6.70 \\ \hline
\textbf{Comp. Sci.} & 6.8 & 6.3 & 5.5 & 5.7 & 5.7 & 6.00 \\ \hline
\textbf{Art} & 5.5 & 6.4 & 5.6 & 5.2 & 5.7 & 5.68 \\ \hline
\textbf{History} & 5.4 & 6.1 & 5.1 & 5.2 & 4.8 & 5.32 \\ \hline
\textbf{Geology} & 5.2 & 5.7 & 5.2 & 4.7 & 5.2 & 5.20 \\ \hline
\textbf{Economics} & 6.5 & 5.7 & 5 & 4.2 & 3.1 & 4.90 \\ \hline
\textbf{Biology} & 5.5 & 5.2 & 4.7 & 4.3 & 4.2 & 4.78 \\ \hline
\textbf{Education} & 5.5 & 4.9 & 3.8 & 4.0 & 2.7 & 4.17 \\ \hline
\textbf{Mean} & \textbf{7.68} & \textbf{7.54} & \textbf{7.06} & \textbf{6.79} & \textbf{6.78} & \\ \hline
\end{tabular}
\caption{Mean citation relevance scores for all domains.}
\label{tab:tabCITE}
\end{minipage}%
\vspace{-20pt}
\end{table}
\vspace{-20pt}
\section{Discussion} \label{discussion}
\vspace{-5pt}
Whilst the overall performance of CORE-GPT is very good, there are still some limitations to consider. CORE-GPT draws its answers and references from the body of Open Access literature. Whilst OA now covers a growing proportion of published scientific articles, there is still a significant quantity that is locked behind publishers' paywalls which CORE-GPT cannot currently access. However this problem, and the issues with current publishing paradigms in general, extend far beyond the scope of this study. 

It should be noted that whilst CORE-GPT was tested across a wide range of domains, only five questions per domain were used for the evaluation. This was to limit the burden on the annotators who validated 100 answers and checked all 500 links to references. Further evaluation could therefore be undertaken with a larger cohort of annotators.  

In the questions dataset, a small number of questions are somewhat basic and not really at the level that would be expected of a research question. Further, it can be seen that there is overlap in the phrasing of some questions, leading to similar questions in some domains. Whilst this reduced the variety of questions by a small margin, we remain confident in the overall results presented here. 

Across all domains there is very strong correlation between the comprehensiveness, trust and utility scores for the answers and the relevance of the citations (Pearson $r = 0.77$ (comp.); $r = 0.83$ (trust); $r = 0.80$ (utility), $n=20$) This indicates that it is access to high quality, relevant literature that is central to delivering high quality answers.




\vspace{-5pt}
\section{Conclusion}
\vspace{-5pt}
In this paper we introduce CORE-GPT a framework that combines LLMs and massive-scale Open Access scientific corpora to deliver a trustworthy, evidence-based question-answering platform. CORE-GPT is an overtly simple, yet elegant solution to the problems that arise when LLMs are asked to provide factual, evidence-based answers. Our evaluation results demonstrate that the answers provided by CORE-GPT are, on the whole, comprehensive, useful and most importantly trustworthy. Further, all references generated by the platform are, by design, genuine research papers held within CORE.

\section{Future Work} \label{future}
\vspace{-5pt}
The results from the evaluation show that CORE-GPT performs well across the majority of scientific domains. This provides a strong foundation to now develop a range of applications using the central CORE-GPT architecture. The initial version of CORE-GPT uses the titles and abstracts of the five most relevant papers as source for the given answers. Due to the limitations in the number of tokens that can be passed to the GPT4 model it is not currently possible to pass the entire full-text content of all papers. This is something that will undoubtedly change in the future and may lead to even stronger results. 

Our initial plan includes making the current version of CORE-GPT available as an addition to the CORE API V3.0. Further, CORE provides a range of management tools for repositories and we see strong potential in developing both an embedded repository version of the service and also a recommender system for repositories based on the CORE-GPT architecture. 

\section{Data and Code Availability}
\vspace{-5pt}
All data and software code used for the evaluation of CORE-GPT are available to promote transparency and reproducibility of the findings. The dataset of questions and answers and the source code used for the analysis and visualisations in this study are accessible on the CORE-GPT GitHub repository\footnote{https://github.com/oacore/core-gpt-evaluation}. Any questions or requests for further information can be addressed to the corresponding author.

\bibliographystyle{vancouver}
\bibliography{main}

\end{document}